\documentclass{article}

\usepackage[preprint,nonatbib]{neurips_2024}

\usepackage[utf8]{inputenc} % allow utf-8 input
\usepackage[T1]{fontenc}    % use 8-bit T1 fonts
\usepackage{url}            % simple URL typesetting
\usepackage{booktabs}       % professional-quality tables
\usepackage{amsfonts}       % blackboard math symbols
\usepackage{nicefrac}       % compact symbols for 1/2, etc.
\usepackage{microtype}      % microtypography
\usepackage{xcolor}         % colors

\usepackage{caption}
\usepackage{subcaption}

\usepackage{graphicx}
\usepackage{amsmath}
\usepackage{wrapfig}
\usepackage{colortbl}
\usepackage{cite}

\usepackage{multirow}
\newcommand{\et}[2]{${#1}^{\pm{#2}}$}
\newcommand{\etb}[2]{$\mathbf{{#1}}^{\pm{#2}}$}

\newcommand{\etal}{\textit{et al}.}
\newcommand{\ie}{\textit{i}.\textit{e}.}
\newcommand{\eg}{\textit{e}.\textit{g}.}

\newcommand{\parsection}[1]{\noindent\textbf{#1:}~}

\definecolor{myblue}{RGB}{0, 112, 192}

\definecolor{citecolor2}{HTML}{0071bc}
\usepackage[pagebackref=true,breaklinks=true,letterpaper=true,colorlinks,urlcolor=citecolor2,citecolor=citecolor2,linkcolor=red,bookmarks=false]{hyperref}

\title{Learning Generalizable Human Motion Generator with Reinforcement Learning}

\author{Yunyao~Mao$^{1,*}$~~~Xiaoyang~Liu$^{1,*}$~~~Wengang~Zhou$^{1,\dagger}$~~~Zhenbo~Lu$^2$~~~Houqiang~Li$^{1,\dagger}$ \\
        {\normalsize $^1$ CAS Key Laboratory of Technology in GIPAS, EEIS Department,} \\
	{\normalsize University of Science and Technology of China} \\
 {\normalsize $^2$ Institute of Artificial Intelligence, Hefei Comprehensive National Science Center} \\
	{\tt\small myy2016@mail.ustc.edu.cn, liuxiaoyang@mail.ustc.edu.cn, zhwg@ustc.edu.cn} \\
    {\tt\small luzhenbo@iai.ustc.edu.cn, lihq@ustc.edu.cn}
}

\begin{document}
\footnotetext[1]{Equal contribution}
\footnotetext[2]{Corresponding authors: Wengang Zhou and Houqiang Li}

\maketitle

\begin{abstract}
Text-driven human motion generation, as one of the vital tasks in computer-aided content creation, has recently attracted increasing attention. While pioneering research has largely focused on improving numerical performance metrics on given datasets, practical applications reveal a common challenge: existing methods often overfit specific motion expressions in the training data, hindering their ability to generalize to novel descriptions like unseen combinations of motions. This limitation restricts their broader applicability. We argue that the aforementioned problem primarily arises from the scarcity of available motion-text pairs, given the many-to-many nature of text-driven motion generation. To tackle this problem, we formulate text-to-motion generation as a Markov decision process and present \textbf{InstructMotion}, which incorporate the trail and error paradigm in reinforcement learning for generalizable human motion generation. Leveraging contrastive pre-trained text and motion encoders, we delve into optimizing reward design to enable InstructMotion to operate effectively on both paired data, enhancing global semantic level text-motion alignment, and synthetic text-only data, facilitating better generalization to novel prompts without the need for ground-truth motion supervision. Extensive experiments on prevalent benchmarks and also our synthesized unpaired dataset demonstrate that the proposed InstructMotion achieves outstanding performance both quantitatively and qualitatively.

\end{abstract}

\section{Introduction}

Human motion generation \cite{zhu2023human,badler1993simulating,aliakbarian2020stochastic,bouazizi2022motionmixer,butepage2017deep,fragkiadaki2015recurrent,barsoum2018hp,mao2019learning,habibie2017recurrent,yan2018mt,aristidou2022rhythm,lee2019dancing} has emerged as a compelling research topic within the realm of computer-aided content generation, attracting substantial scholarly interest in recent years. This task holds significant promise across a spectrum of applications, spanning from film character animation, gaming to interactive storytelling in virtual or augmented reality. The ability to synthesize realistic and contextually appropriate human motions from textual descriptions represents a crucial step towards creating immersive and engaging digital experiences.

\begin{figure}[t]
  \centering
  \includegraphics[width=1.0\textwidth]{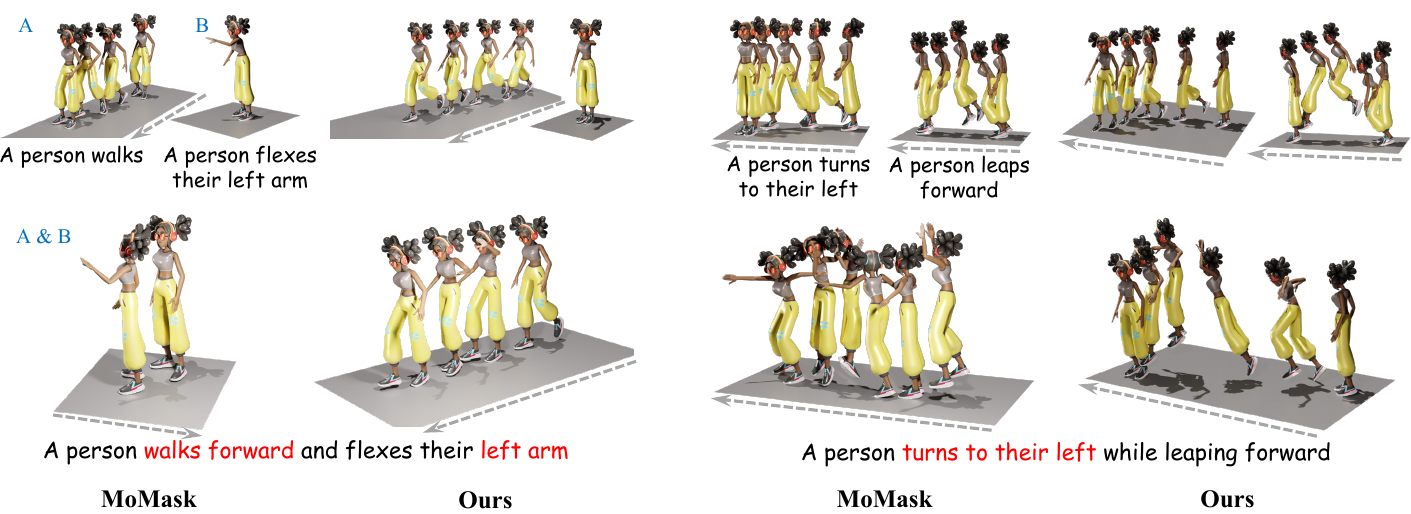}
  \caption{Examples generated from simple and compositional given textual descriptions. Our method significantly outperforms previous state-of-the-art method MoMask \cite{momask} in terms of generalization capability to novel motion compositions. The compositional descriptions are generated with the aid of large language models, as discussed in Section \ref{data_pipeline}.}
  \label{fig:intro}
\end{figure}

Text-to-motion generation \cite{ahn2018text2action} is a challenging cross-modal task owing to the diversity of textual descriptions and human motions. Algorithms need to understand the rich contextual relationships embedded in the textual descriptions and translate them into precise dynamic motion sequences. To this end, many early studies have sought to align the representation of motion sequences with textual descriptions within prevalent generative frameworks like auto-encoders \cite{language2pose,ghosh2021synthesis,motionclip} and variational auto-encoders \cite{petrovich2021action,petrovich2022temos}. More recently, methods based on generative transformers \cite{tm2t,t2mgpt,momask,motiongpt,zhang2024motiongpt,lu2023humantomato} and diffusion models \cite{mdm,motiondiffuse,remodiffuse,mofusion,physdiff,amd,jin2024act,wang2023fg,hu2023motion} are proposed for generating human motion conditioned on more complex textual descriptions.

While numerical performance metrics on specific datasets are improved with advances in algorithm and model design, the generalization capability of existing text-to-motion generators remain underexplored. Empirical observations reveal that recent top-performing models tend to overfit to training samples due to the scarcity of paired training data \cite{kitml,humanml3d}. As illustrated in Figure \ref{fig:intro}, when users provide prompts describing novel combinations of simple motions seen during training, the model encounters difficulties in accurately generating the corresponding motion sequences. The limited generalization capability induces significant challenges for subsequent applications.

An intuitive solution to enhance the generalization capability involves expanding the dataset by gathering more motion-text pairs for training. Nevertheless, unlike the training of discriminative models, the many-to-many nature of the text-to-motion generation necessitates a notably higher volume of training samples. Moreover, acquiring reliable human motion sequences involves volunteers wearing motion capture devices to perform predetermined motions within controlled environments, resulting in time-consuming, labor-intensive, and costly data collection procedures.

To this end, we try to tackle the aforementioned problem in an alternative way. We present \textbf{InstructMotion}, a reinforcement learning-based framework for generalizable human motion generation. Given a supervised pre-trained text-to-motion generator, we further \textit{instruct} it to align the output motion with the given textual descriptions. This approach draws inspiration from the recent advancements in the reinforcement learning with human feedback (RLHF) \cite{gpt4} applied in training large language models. RLHF is designed to better align the outputs of large language models to the complex human preferences. In RLHF, a reward model is first trained using human annotated samples. It is expected to provide reliable human preference scores for novel responses given by LLMs, thereby facilitating subsequent reinforcement learning-based fine-tuning.

Similarly, we expect a reward model that is capable of assessing the \textit{alignment} between the generated motion sequences and the given textual descriptions. In this way, InstructMotion can work on both previously seen prompts for better global semantic level alignment and unseen ones for better generalization capabilities, without the need for ground-truth motion sequences. The previously unseen textual descriptions can be easily generated by large language models. Unlike the reward model learning paradigm in typical RLHF schemes that require additional human preference labeling, we opt to design the computation of reward values based on contrastive pre-trained text and motion encoders. Specifically, we adopt the negative Euclidean distance between the global motion and text embedding as the metric for assessing their semantic alignment. Given the great success of RLHF in language models, in our approach, we adopt the autoregressive transformer-based motion generator as the baseline approach. In this paradigm, the next motion token prediction can be naturally interpreted as a Markov decision process (MDP), as required by reinforcement learning frameworks.

To verify the effectiveness of the proposed approach, we conduct extensive experiments on prevalent benchmarks like HumanML3D \cite{humanml3d} and KIT-ML \cite{kitml} and also our generated unpaired dataset with novel motion descriptions. Results show that after fine-tuning with our InstructMotion framework, the text-to-motion generator exhibits superior performance both quantitatively and qualitatively. Notably, under human evaluation, InstructMotion significantly surpasses previous state-of-the-art methods in generating novel motion sequences.

\section{Related Work}
\parsection{Text-driven Human Motion Generation}
Over the past few years, significant strides have been observed in the field of text-driven human motion generation. In many earlier works, the representation of motion and text are aligned in the same embedding space. For example, Language2Pose \cite{language2pose} learns the joint embedding with a curriculum learning strategy. In this way, the text embedding can be fed into the motion decoder to generate corresponding motion sequence during inference. Similar idea can be found in Ghosh \etal \cite{ghosh2021synthesis}. and MotionCLIP \cite{motionclip}. The former learns separate embeddings for different parts of body movements and the latter learns the motion representation under the guidance of the text and image embeddings extracted by CLIP \cite{clip} encoders. 

The temporal VAE framework is firstly introduced in \cite{action2video,petrovich2021action} for class-conditional motion synthesis. After that, T2M \cite{humanml3d} and TEMOS \cite{petrovich2022temos} extend this idea for text-to-motion generation. TEACH \cite{teach} further extends TEMOS in an autoregressive manner, where long motion sequences can be generated under the guidance of successive textual descriptions.

Inspired by the recent success of encoder-decoder and decoder-only architecture in natural language processing, TM2T \cite{tm2t} and T2M-GPT \cite{t2mgpt} are successively introduced. They both learn a vector quantized variational autoencoder to convert the continuous motion sequences into discrete motion tokens. TM2T adopts the encoder-decoder architecture that treats text-driven motion generation and motion comprehension as bidirectional translation tasks. Differently, T2M-GPT utilizes a transformer decoder to perform next motion token prediction, with the CLIP text embedding serving as the \texttt{<start of sentence>} token.

Given the great success of the diffusion model \cite{ddpm} in natural image generation, MDM \cite{mdm} and MotionDiffuse \cite{motiondiffuse} incorporate it for text-to-motion generation. ReMoDiffuse \cite{remodiffuse} extends MotionDiffuse by integrating a retrieval mechanism to enhance the denoising process, yielding a retrieval-augmented motion diffusion model.

\parsection{Reinforcement Learning with Human Feedback}
The advent of powerful large language models (LLMs) like ChatGPT has revolutionized the field of Natural Language Processing (NLP). A critical aspect of developing such models involves aligning their outputs with human preferences to ensure ethical, relevant, and coherent responses. Reinforcement learning with human feedback (RLHF) has emerged as a pivotal methodology in this pursuit. Initial explorations into RLHF began with researches that seek to incorporate human judgments into model optimization. For example, Christiano \etal \cite{christiano2017deep} proposes an approach where human evaluators rank model responses, and these rankings are used to shape the reward function in simulated robotic control. Similar ideas can also be found in earlier natural language processing tasks like machine translation \cite{nguyen2017reinforcement}, question-answering \cite{nakano2021webgpt}, summarization \cite{stiennon2020learning,ziegler2019fine}, and instruction-following \cite{ouyang2022training}.

According to the technical report from OpenAI \cite{gpt4}, the pre-trained GPT model is firstly fine-tuned using a dataset of human-labeled responses to various prompts. This step helps the model understand and mimic how humans might respond in different contexts. Next, human evaluators rate the model's responses on a scale based on factors such as relevance, coherence, ethics, and overall quality. These ratings serve as the ground truth for training a separate \textit{reward model} which learns to predict how well a given response will be rated by humans. The language model interacts with the environment (novel prompts), and its actions (responses) are scored by the reward model. High-scoring responses receive positive reinforcement, adjusting the model's parameters to optimize for generating similar high-quality and human-preferred outputs.

Our approach takes inspiration from the RLHF algorithm. Differently, we do not use human feedback but utilize contrastive pre-trained text and motion encoders to design our reward model, which is capable of assessing the alignment between the generated motion and the given textual descriptions.

\section{Preliminary}

\subsection{Autoregressive Transformer-based Motion Generator}
\label{t2m_gpt}
In T2M-GPT \cite{t2mgpt}, the authors present a novel approach for generating human motion from textual descriptions using a combination of vector quantized variational autoencoder (VQ-VAE) \cite{vqvae} and generative pre-trained transformer (GPT) \cite{radford2018improving,vaswani2017attention}. Specifically, the VQ-VAE is trained using a CNN-based architecture with commonly used training techniques such as exponential moving average (EMA) and code reset. It maps the input motion sequence $\mathbf{m} = \{x_1, x_2, \cdots, x_T\}$ into a sequence of indices $\mathbf{s} = \{s_1, s_2, \cdots, s_{T/l}\}$ with  temporal downsampling rate $l$, where $s_i \in [1, 2, \cdots, K]$ are the index of the vector in the learned codebook $C = \{c_k\}_{k=1}^{K}$ with $c_k \in R^{d_c}$.

After that, text-to-motion generation is formulated as an next-index prediction task performed by an autoregressive transformer. Specifically, the distribution of index $s_t$ is predicted conditioned on the text embedding $c$ and the previously generated sequence of indices up to $t-1$, represented as $s_{<t}$. This can be mathematically expressed as $p(s_t | c, s_{<t})$. The text embedding serves as the \texttt{start-of-sentence} token and is extracted using the CLIP \cite{clip} text encoder, and an \texttt{end-of-sentence} token is added to the sequence of indices, signifying the end of motion generation. 

The introduction of autoregressive transformer-based text-to-motion models has enabled us to represent text-driven motion generation as a Markov decision process, akin to the practices in large language models, thereby facilitating reinforcement learning-based fine-tuning.

\begin{figure}[t]
  \centering
  \includegraphics[width=1.0\textwidth]{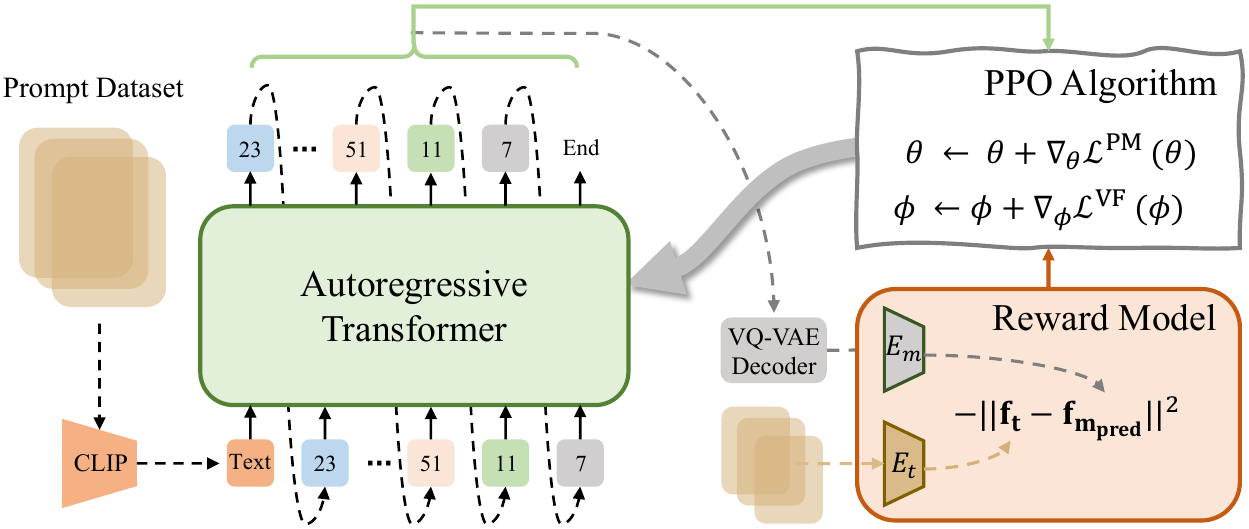}
  \caption{The over pipeline of InstructMotion. Given a batch of textual prompts, the pre-trained autoregressive generator first produces the corresponding motion sequences, which is fed into the reward model along with the text prompts to assess the generation quality, yielding a matching score. The score, combined with the prediction logits and the the critic model output (omitted in the figure), is then organized by the PPO algorithm to optimize the generator in a inner training loop.}
  \label{fig:method}
\end{figure}

\section{Methodology}

Commencing with a conventionally supervised-trained text-to-motion model that commonly struggles with generalization due to limited paired training data available, our proposed InstructMotion innovatively addresses this shortcoming by incorporating the reinforcement learning (RL) strategy. This allows for the fine-tuning of the pre-trained model even without direct access to ground-truth motion sequences corresponding to the textual inputs.

There are two prerequisites to facilitate reinforcement learning training. The first entails formulating text-driven motion generation as a Markov Decision Process (MDP). This goal has been met, as discussed in Section~\ref{t2m_gpt}. The second necessitates defining an appropriate reward function for the model's responses to textual prompts, \ie, the generated motion sequences. We will elaborate on the definition of the reward model used in our approach in Section~\ref{reward_design}. To construct unpaired (motion descriptions only) data for reinforcement learning training, we further design a motion description generation pipeline with the aid of large language model. It is capable of producing a substantial amount of novel textual descriptions, derived from fundamental and recurring motion patterns (termed meta-motions) extracted from the original dataset. We will go into detail in Section~\ref{data_pipeline}. Finally, the adopted reinforcement learning training procedure is introduced in Section~\ref{train_procedure}.

\subsection{Reward Design}
\label{reward_design}
To construct the reward model, an intuitive solution is to refer the practices in the training of the large language model, \ie, training with human annotated preference data. However, collecting a large amount of labeled data requires significant labor costs, which is contrary to our original intent. To this end, we propose to design the reward function using contrastive pre-trained motion and text encoders. Specifically, we adopt the motion and text encoders in \cite{humanml3d}, which are trained with the following objective:
\begin{equation}
\mathcal{L}_{CL} = (1 - y) (\| \mathbf{f_t} - \mathbf{f_m}\|)^2 + y (max(0, m - \| \mathbf{f_t} - \mathbf{f_m}\|))^2,
\end{equation}
where $y$ is a binary label indicating the matched text-motion pairs and $m$ is the manually set margin. $\mathbf{f_t} = E_t(\mathbf{t})$ and $\mathbf{f_m} = E_m(\mathbf{m})$ are embeddings extracted by text and motion encoders, respectively.

$E_t$ and $E_m$ map the text and motion modality into a shared embedding space, where semantically similar textual descriptions and motion sequences exhibit smaller Euclidean distances. Given a textual description $\mathbf{t}$ and the generated motion $\mathbf{m}_{\mathrm{pred}}$ conditioned on $\mathbf{t}$, we can thus design the reward function as follows:
\begin{equation}
    \label{eq:motion_padding}
    r (\mathbf{t}, \mathbf{m}) = \left\{
    \begin{array}{ll}
    - \lambda_t \| \mathbf{f}_{\mathbf{t}} - \mathbf{f}_{\mathbf{m}_{\mathrm{pred}}}\|^2, & {\text{text-only}} \\
    - \lambda_t \| \mathbf{f}_{\mathbf{t}} - \mathbf{f}_{\mathbf{m}_{\mathrm{pred}}}\|^2 - \lambda_m \| \mathbf{f}_{\mathbf{m}_{\mathrm{gt}}} - \mathbf{f}_{\mathbf{m}_{\mathrm{pred}}}\|^2, & \text{paired}
    \end{array}
    \right.
    ,
\end{equation}
where $\mathbf{m}_{\mathrm{gt}}$ is the ground-truth motion sequence. $\lambda_t$ and $\lambda_m$ are weighting factors. For paired training data, we also experiment with incorporating the squared Euclidean distance metric between the global representation of the ground-truth motion sequence and the predicted one in reward computation.

\setlength{\intextsep}{0pt}
\begin{wrapfigure}[11]{R}{0.45\textwidth}
	\centering
	\includegraphics[width=1.0\linewidth]{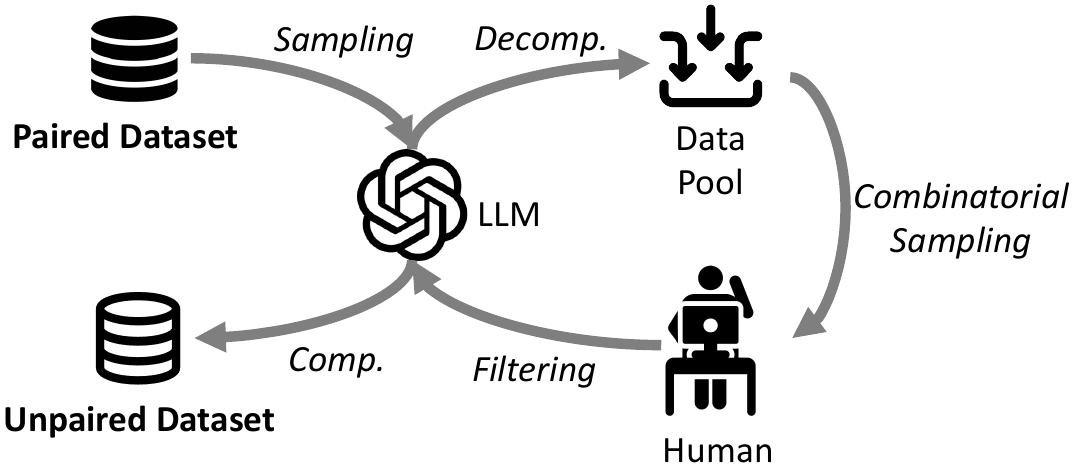}
	\caption{An illustration of the LLM-assisted novel motion description generation pipeline.}\label{fig:data_pipeline}
\end{wrapfigure}

\subsection{Novel Motion Description Generation}
\label{data_pipeline}

In the data processing pipeline, we begin by randomly sampling motion descriptions from the original dataset. Next, we use an large language model to determine whether each sampled description can be decomposed. Descriptions that cannot be decomposed are directly added to the data pool. Conversely, the ones that can be decomposed are further split into meta motions using the language model, and the obtained meta motions are then added to the pool. To illustrate with an example, consider the following motion descriptions: "a person walks forward, then stutters on his right leg." and "a person flexes their left arm." The first description is decomposable, and thus split into "a person walks forward." and "a person stutters on his right leg." The second description is indivisible and is added directly as "a person flexes their left arm." As a result, the meta motion dataset now includes: 1 ("a person walks forward."), 2 ("a person stutters on his right leg."), and 3 ("a person flexes their left arm.").

After that, we can then randomly combine meta motions and utilize the large language model to compose novel motion descriptions. For instance, if meta motions 1 ("a person walks forward.") and 3 ("a person flexes their left arm.") are combined, the new synthesized motion description would be "a person walks forward and flexes their left arm." Note that this combination process is overseen by human evaluators, ensuring that implausible motion descriptions are filtered out. Finally, we perform part-of-speech (POS) tagging on the resulting motion descriptions to generate the synthesized unpaired dataset.

\subsection{Reinforcement Learning Training Procedure}
\label{train_procedure}

Akin to practices adopted in advanced language models, we integrate reinforcement learning techniques to facilitate semantic alignment between the generated motion sequences and the provided textual descriptions, which entails maximizing the following objective:
\begin{equation}
    \mathcal{J}_{r}(\pi_{\theta})=\mathbb{E}_{\mathbf{t}\sim p_\mathrm{data},\mathbf{m}\sim\pi_\theta} \left[{r(\mathbf{t},\mathbf{m})-\beta\log\frac{\pi_\theta(\mathbf{m}\mid\mathbf{t})}{\pi_\mathrm{ref}(\mathbf{m}\mid\mathbf{t})}}\right],
\end{equation}
where $r$ is the reward function measuring the degree of the alignment between textual prompt $\mathbf{t}$ and the corresponding generated motion sequence $\mathbf{m}$. $\pi_\mathrm{ref}$ is the reference model, essentially a fixed copy of the pre-trained text-to-motion model that serves to regulate the learning process of the actor model $\pi_{\theta}$ (with an intensity factor $\beta$). This prevents the policy from moving too far away from the distribution where reward function is valid.

As shown in Figure~\ref{fig:method}, in our approach, we adopt standard reinforcement learning method: proximal policy optimization \cite{ppo} (PPO) as the training algorithm. PPO adopts the actor-critic framework. The actor network learns the policy $\pi_{\theta}$, parameterized by $\theta$, which maps states to probabilities of taking different actions. Specifically in motion generation, the actor network corresponds to the text-to-motion transformer to be fine-tuned and the action pertains to the selection of motion tokens in current time step. The critic network estimates the state-value function $V_{\phi}(s)$, giving an estimate of how good it is for the agent to be in a particular state. To be computationally efficient, the actor and critic network in our approach share the first few layers and the detach operation ensures that the learning of the critical network does not affect the shared part.

Based on the final reward and the state-value, the return $G$ and advantage $A$ can be computed accordingly. We refer to the details of computation to the original paper of PPO \cite{ppo}. The loss function for optimizing actor network is as follows:
\begin{equation}
\mathcal{L}^{\mathrm{PM}}(\theta) = \mathbb{E}_{t}\left[\min\left(\frac{\pi_{\theta}(a_t | s_t)}{\pi_{\theta_{\mathrm{old}}}(a_t | s_t)} A_t, \text{clip}\left(\frac{\pi_{\theta}(a_t | s_t)}{\pi_{\theta_{\mathrm{old}}}(a_t | s_t)}, 1-\epsilon, 1+\epsilon\right) A_t\right)\right],
\end{equation}
where $\theta$ and $\theta_{\mathrm{old}}$ denote the parameters of the new and old policies (\ie, the text-to-motion generator in our approach), respectively. $\pi_{\theta}(a_t | s_t)$ and $\pi_{\theta_{\mathrm{old}}}(a_t | s_t)$ are the probability of selecting action $a_t$ in state $s_t$ under the new and old policy, respectively. Here in auto-regressive transformer-based text-to-motion generator, they refer to the probability of selecting current motion token $p(s_t | c, s_{<t})$. $A_t$ is the estimated advantage at current time step $t$. $\epsilon$ is the clipping threshold that ensures the improved policy stays close to its previous version.

The critic network is optimized to estimate the return value $G_t$ at current time step $t$:
\begin{equation}
\mathcal{L}^{\mathrm{VF}}(\phi) = \mathbb{E}_{t}\left[\left((V_{\phi}(s_t) - G_t\right)^2\right].
\end{equation}

\section{Experiments}

\subsection{Implementation Details}
We choose T2M-GPT as our baseline approach. The text-to-motion transformer consists of 18 layers, each characterized by a dimension of 1,024 and 16 heads. We configure the PPO algorithm with a mini-batch size of 128, aligning it with the outer loop batch size. The number of epochs for PPO is set to 2. The AdamW \cite{adamw} optimizer is adopted with $\beta_1$=0.9, $\beta_2$ = 0.99. We fine-tune the pre-trained text-to-motion generator for 40k iterations with a learning rate of 5e-6. All the experiments are conducted on 4 NVIDIA RTX 3090 GPUs.

\subsection{Comparison with the State-of-the-art Methods}

\begin{table}[t]
    \setlength{\tabcolsep}{4pt}
    \caption{\textbf{Quantitative comparison on HumanML3D~\cite{humanml3d} test set.} The evaluation metrics are computed following Guo \etal~\cite{humanml3d}. The evaluation is repeated 20 times for confidence interval estimation. $^\S$ A indicates reliance on ground-truth sequence length for generation.}
    \label{table:humanml3d}
    \centering
    \small
    \begin{tabular}{l c c c c c c}
    \toprule
    \multirow{2}{*}{Methods}  & \multicolumn{3}{c}{R-Precision $\uparrow$} & \multirow{2}{*}{FID $\downarrow$} & \multirow{2}{*}{MM-Dist $\downarrow$} & \multirow{2}{*}{Diversity $\uparrow$} \\

    \cline{2-4}
    ~ & Top-1 & Top-2 & Top-3 \\
    
    \midrule
        TEMOS$^\S$~\cite{petrovich2022temos} & \et{0.424}{.002} & \et{0.612}{.002} & \et{0.722}{.002} & \et{3.734}{.028} & \et{3.703}{.008} & \et{8.973}{.071} \\
        MLD$^\S$~\cite{mld} & \et{0.481}{.003} & \et{0.673}{.003} & \et{0.772}{.002} & \et{0.473}{.013} & \et{3.196}{.010} & \etb{9.724}{.082} \\
        MDM$^\S$~\cite{mdm} & - & - & \et{0.611}{.007} & \et{0.544}{.044} & \et{5.566}{.027} & \et{9.559}{.086} \\ 
        MotionDiffuse$^\S$~\cite{motiondiffuse} & \et{0.491}{.001} & \et{0.681}{.001} & \et{0.782}{.001} & \et{0.630}{.001} & \et{3.113}{.001} & \et{9.410}{.049} \\
        GraphMotion$^\S$~\cite{jin2024act} & \et{0.504}{.003} & \et{0.699}{.002} & \et{0.785}{.002} & \et{0.116}{.007} & \et{3.070}{.008} & \et{9.692}{.067} \\
        ReMoDiffuse$^\S$~\cite{remodiffuse} & \et{0.510}{.005} & \et{0.698}{.006} & \et{0.795}{.004} & \et{0.103}{.004} & \et{2.974}{.016} & \et{9.018}{.075} \\
        MoMask$^\S$~\cite{momask} & \etb{0.521}{.002} & \etb{0.713}{.002} & \etb{0.807}{.002} & \etb{0.045}{.002} & \etb{2.958}{.008} & - \\
        \midrule
        Language2Pose~\cite{language2pose} & \et{0.246}{.002} & \et{0.387}{.002} & \et{0.486}{.002} & \et{11.02}{.046} & \et{5.296}{.008} & \\
        Ghosh \etal~\cite{ghosh2021synthesis} & \et{0.301}{.002} & \et{0.425}{.002} & \et{0.552}{.004} & \et{6.532}{.024} & \et{5.012}{.018} \\
        TM2T~\cite{tm2t} & \et{0.424}{.003} & \et{0.618}{.003} & \et{0.729}{.002} & \et{1.501}{.017} & \et{3.467}{.011} & \et{8.589}{.076} \\
        Guo \etal~\cite{humanml3d} & \et{0.455}{.003} & \et{0.636}{.003} & \et{0.736}{.002} & \et{1.087}{.021} & \et{3.347}{.008} & \et{9.175}{.083} \\ 
        T2M-GPT \cite{t2mgpt} & \et{0.491}{.003} & \et{0.680}{.003} & \et{0.775}{.002} & \et{0.116}{.004} & \et{3.118}{.011} & \etb{9.761}{.081} \\
        Fg-T2M \cite{wang2023fg} & \et{0.492}{.002} & \et{0.683}{.003} & \et{0.783}{.002} & \et{0.243}{.019} & \et{3.109}{.007} & \et{9.278}{.072} \\
        MotionGPT \cite{motiongpt} & \et{0.492}{.003} & \et{0.681}{.003} & \et{0.778}{.002} & \et{0.232}{.008} & \et{3.096}{.008} & \et{9.528}{.071} \\
        \textbf{Ours} & \etb{0.505}{.003} & \etb{0.694}{.003} & \etb{0.790}{.003} & \etb{0.099}{.003} & \etb{3.028}{.011} & \et{9.741}{.093} \\
    \bottomrule
    \end{tabular}
\end{table}
\begin{table}[t]
    \setlength{\tabcolsep}{4pt}
    \caption{\textbf{Quantitative comparison on KIT-ML~\cite{kitml} test set.} The evaluation metrics are computed following Guo \etal~\cite{humanml3d}. The evaluation is repeated 20 times for confidence interval estimation.}
    \label{table:kitml}
    \centering
    \small

    \begin{tabular}{l c c c c c c}
    \toprule
    \multirow{2}{*}{Methods}  & \multicolumn{3}{c}{R-Precision $\uparrow$} & \multirow{2}{*}{FID $\downarrow$} & \multirow{2}{*}{MM-Dist $\downarrow$} & \multirow{2}{*}{Diversity $\uparrow$} \\

    \cline{2-4}
    ~ & Top-1 & Top-2 & Top-3 \\
    
    \midrule
        TEMOS$^\S$~\cite{petrovich2022temos} & \et{0.353}{.002} & \et{0.561}{.002} & \et{0.687}{.002} & \et{3.717}{.028} & \et{3.417}{.008} & \et{10.84}{.071} \\
        MLD$^\S$~\cite{mld} & \et{0.390}{.008} & \et{0.609}{.008} & \et{0.734}{.007} & \et{0.404}{.027} & \et{3.204}{.027} & \et{10.80}{.117} \\
        MDM$^\S$~\cite{mdm} & - & - & \et{0.396}{.004} & \et{0.497}{.021} & \et{9.191}{.022} & \et{10.85}{.109} \\ 
        MotionDiffuse$^\S$~\cite{motiondiffuse} & \et{0.417}{.004} & \et{0.621}{.004} & \et{0.739}{.004} & \et{1.954}{.062} & \et{2.958}{.005} & \et{11.10}{.143} \\
        GraphMotion$^\S$~\cite{jin2024act} & \et{0.429}{.007} & \et{0.648}{.006} & \et{0.769}{.006} & \et{0.313}{.013} & \et{3.076}{.022} & \etb{11.12}{.135} \\
        ReMoDiffuse$^\S$~\cite{remodiffuse} & \et{0.427}{.014} & \et{0.641}{.004} & \et{0.765}{.055} & \etb{0.155}{.006} & \et{2.814}{.012} & \et{10.80}{.105} \\
        MoMask$^\S$~\cite{momask} & \etb{0.433}{.007} & \etb{0.656}{.005} & \etb{0.781}{.005} & \et{0.204}{.011} & \etb{2.779}{.022} & - \\
        \midrule
        Language2Pose~\cite{language2pose} & \et{0.221}{.005} & \et{0.373}{.004} & \et{0.483}{.005} & \et{6.545}{.072} & \et{5.147}{.030} \\
        Ghosh \etal~\cite{ghosh2021synthesis} & \et{0.255}{.006} & \et{0.432}{.007} & \et{0.531}{.007} & \et{5.203}{.107} & \et{4.986}{.027} \\
        TM2T~\cite{tm2t} & \et{0.280}{.006} & \et{0.463}{.007} & \et{0.587}{.005} & \et{3.599}{.051} & \et{4.591}{.019} & \et{9.473}{.100} \\
        Guo \textit{et al.}~\cite{humanml3d} & \et{0.361}{.006} & \et{0.559}{.007} & \et{0.681}{.007} & \et{3.022}{.107} & \et{3.488}{.028} & \et{10.72}{.145} \\
        T2M-GPT \cite{t2mgpt} & \et{0.416}{.006} & \et{0.627}{.006} & \et{0.745}{.006} & \et{0.514}{.029} & \et{3.007}{.023} & \et{10.92}{.108} \\
        Fg-T2M \cite{wang2023fg} & \et{0.418}{.005} & \et{0.626}{.004} & \et{0.745}{.004} & \et{0.571}{.047} & \et{3.114}{.015} & \et{10.93}{.083} \\
        \textbf{Ours} & \etb{0.440}{.007} & \etb{0.653}{.007} & \etb{0.774}{.005} & \etb{0.381}{.027} & \etb{2.815}{.018} & \etb{10.95}{.099} \\
    \bottomrule
    \end{tabular}
\end{table}

\parsection{Quantitative Comparisons}
Tables \ref{table:humanml3d} and Table \ref{table:kitml} present an evaluation of our proposed InstructMotion framework across two prevalent datasets: HumanML3D \cite{humanml3d} and KIT-ML \cite{kitml}. Results show that InstructMotion excels in quantitative measures, notably R-Precision and FID \cite{heusel2017gans} scores, surpassing the T2M-GPT \cite{t2mgpt} baseline by considerable margins. This reveals that the motion sequences our model generates are better aligned to the given textual descriptions. Most importantly, while quantitative excellence is evident, the true prowess of InstructMotion lies in its enhanced generalization capability of adeptly transforming novel textual descriptions into appropriate motion sequences, which we will discuss in detail in the following user study and qualitative comparisons.

\begin{figure}[t]
  \centering
  \includegraphics[width=1.0\textwidth]{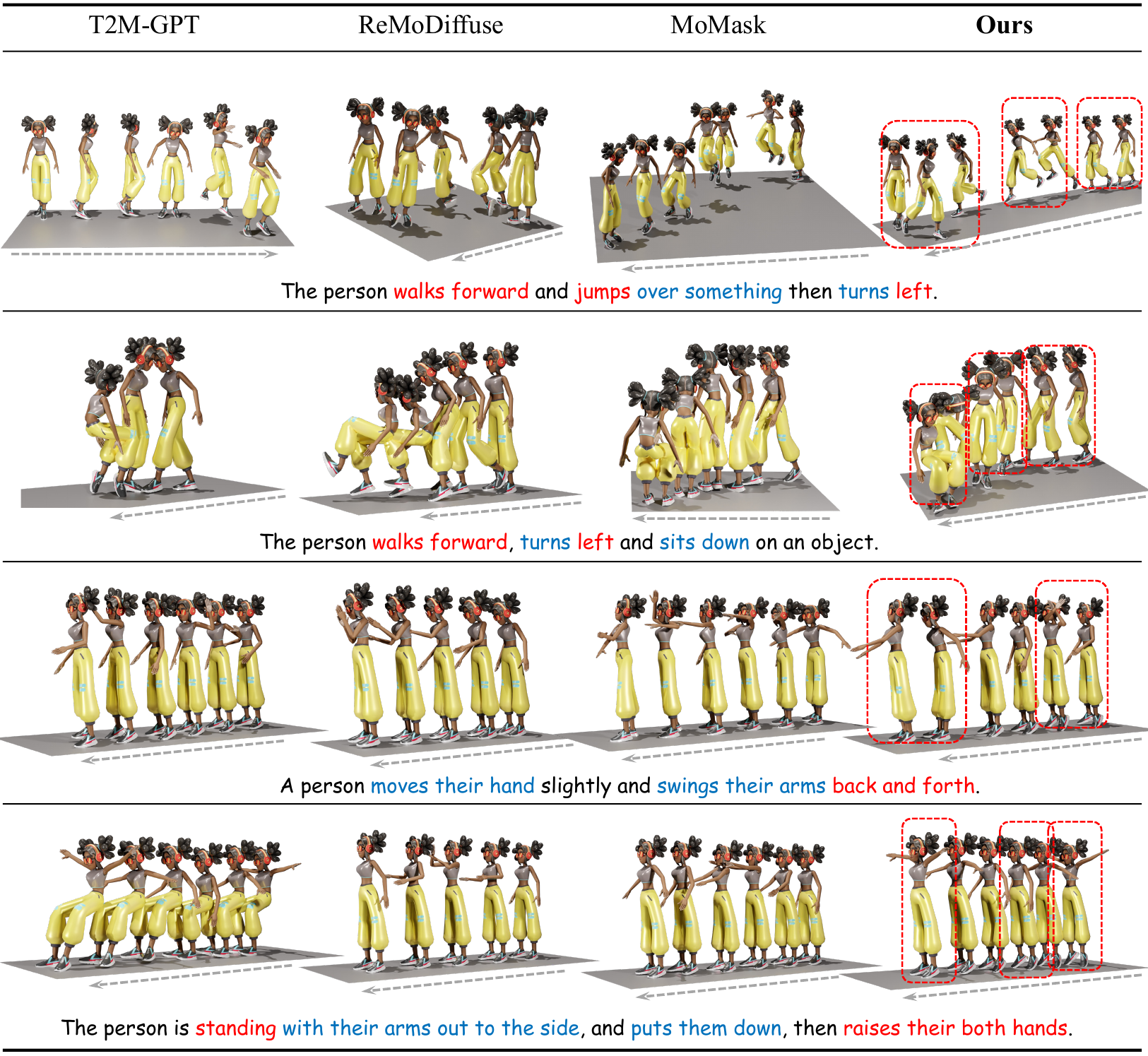}
  \caption{\textbf{Qualitative comparisons with top-performing methods.} Our InstructMotion exhibits enhanced generalization capability and accurately interpret novel combinations of motion instructions.}
  \label{fig:visualization}
\end{figure}

\begin{figure}[t]
\centering
\begin{minipage}{0.435\textwidth}
    \centering
    \includegraphics[width=\textwidth]{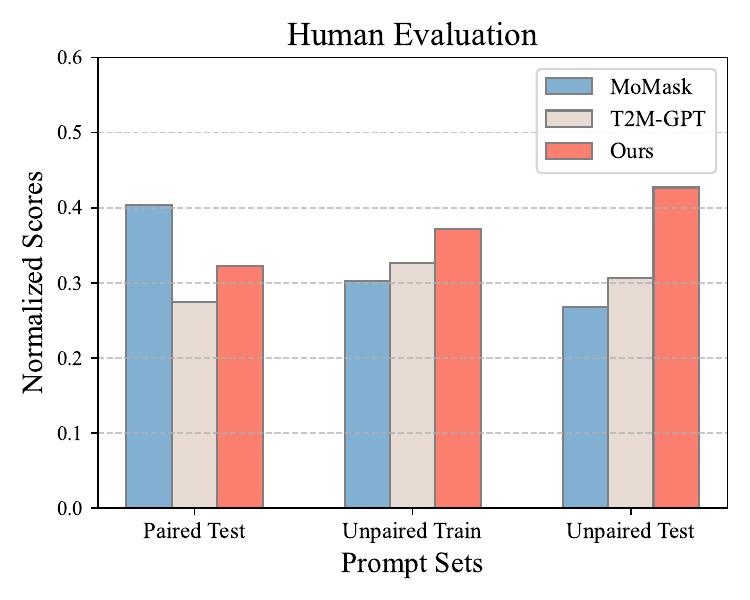}
    \subcaption[]{}
    \label{fig:human}
\end{minipage}
\hspace{\fill}
\begin{minipage}{0.25\textwidth}
    \centering
    \includegraphics[width=\textwidth]{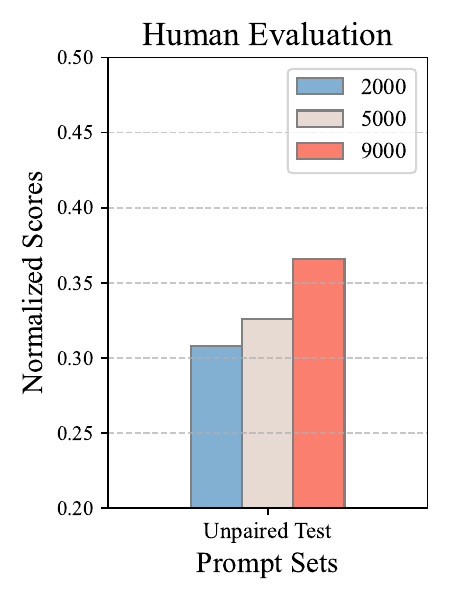}
    \subcaption[]{}
    \label{fig:human_data}
\end{minipage}
\hspace{\fill}
\begin{minipage}{0.25\textwidth}
    \centering
    \includegraphics[width=\textwidth]{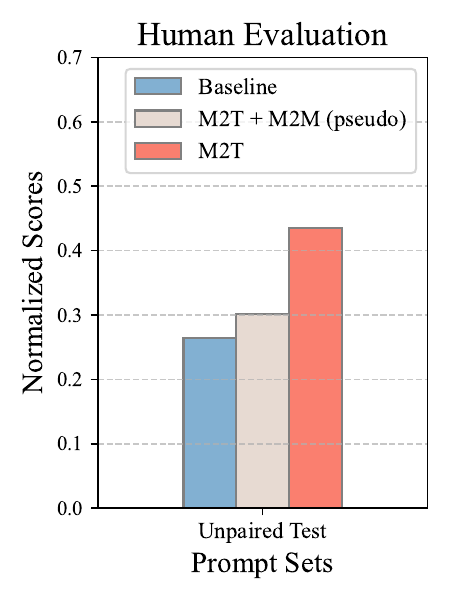}
    \subcaption[]{}
    \label{fig:human_ablation}
\end{minipage}
\caption{\textbf{Human evaluation.} (a) Comparison with the state-of-the-art method MoMask \cite{momask} and the baseline approach T2M-GPT \cite{t2mgpt}. Three different prompt sets are used. (b) Experiment of data scaling. Prompts are sampled from the unpaired test set. (c) Ablative experiment of reward design. Prompts are sampled from the unpaired test set.}
\label{fig:user_study}
\vspace{-2pt}
\end{figure}

\parsection{User Study}
We compare InstructMotion with both the previous top-performing method, MoMask \cite{momask}, and the adopted baseline model, T2M-GPT \cite{t2mgpt}. To ensure a comprehensive assessment, we randomly selected 30 prompts each from the test subset of the HumanML3D dataset (Paired Test), the unpaired synthetic textual prompts used for training (Unpaired Train), and the retained synthetic test data (Unpaired Test), respectively. The results on the HumanML3D test set show that MoMask attained the highest human preference, with a normalized score of 0.40, aligning with the quantitative evaluation metrics reported in Table \ref{table:humanml3d}. Our InstructMotion also performs favorably against the baseline approach T2M-GPT. Since the synthesized dataset contains only the textual descriptions, the PPO fine-tuning process inherently operates in an unsupervised manner. Consequently, we conducted human evaluation on both the training and test subsets of the synthesized dataset. As shown in Figure \ref{fig:human}, our approach exhibits the highest normalized scores (0.37 and 0.43, respectively), outperforming MoMask and T2M-GPT by considerable margins. This suggests that InstructMotion possesses strong generalization capability to novel motion descriptions. Then, we investigate the effects of increasing the scale of synthetic data used during training, as illustrated in Figure \ref{fig:human_data}. Our findings reveal that a greater amount of synthetic data consistently enhances the model's generalization capability. It's worth mentioning that the unsupervised nature also facilitates customized optimization of the model to cater precisely to the motion descriptions outlined by users, a feature that holds significant promise for future exploration and application.

\parsection{Qualitative Comparisons}
Figure \ref{fig:visualization} presents a qualitative evaluation comparing our InstructMotion to T2M-GPT \cite{t2mgpt}, ReMoDiffuse \cite{remodiffuse}, and MoMask \cite{momask}. We can find that although previous methods perform well in terms of quantitative metrics, it still struggle when encountering descriptions with novel and complex motion compositions. For example, MoMask overlooks the expression "walk forward" and both ReMoDiffuse and T2M-GPT fail to include the motion "swinging arms back and forth." Furthermore, T2M-GPT misinterprets "standing," generating it as "sitting" instead. In contrast, our approach exhibits significantly better generalization capability.

\subsection{Ablative Analysis}

\parsection{Reward Design}
As stated in Section \ref{reward_design}, the squared Euclidean distance is adopted as our semantic similarity metric, which can measure both motion-to-text and motion-to-motion alignment during our reward design. As shown in Table \ref{tab:reward_design}, we report the numerical performance on paired dataset HumanML3D. Relying solely on motion-to-text alignment for training improves R-precision but at the cost of increased FID. To this end, we try to integrate motion-to-motion alignment into our strategy, comparing the global features of synthesized sequences with their ground-truth counterparts. Results show that such compositional reward design results in a steady performance improvement over the baseline. In our approach, we assign weights of 0.4 to motion-to-text and 2.0 to motion-to-motion when computing quantitative metrics for paired datasets.

\begin{table}[t]
  \caption{Ablative experiments evaluated on HumanML3D test set. M2T and M2M denote motion-to-text and motion-to-motion alignment metrics, respectively.}
  \vspace{-2pt}
  \small
  \begin{minipage}{.3\textwidth}
    \setlength\tabcolsep{3pt}
    \subcaption{Reward Design}
    \vspace{-1.5mm}
    \label{tab:reward_design}
    \centering
    \begin{tabular}{c c c c}
        \toprule[1.0pt]
        M2T & M2M & FID $\downarrow$ & Top-1 $\uparrow$ \\
        \hline
        2.0 & / & 0.127 & \textbf{0.514} \\
        / & 2.0 & 0.112 & 0.499 \\
        0.4 & 2.0 & \textbf{0.099} & 0.505 \\
        0.2 & 2.0 & 0.104 & 0.503 \\
        0.1 & 2.0 & 0.109 & 0.502 \\
        \bottomrule[1.0pt]
    \end{tabular}
  \end{minipage}
  \hspace{.012\textwidth}
  \begin{minipage}{.34\textwidth}
    \setlength\tabcolsep{3pt}
    \subcaption{Critic Network}
    \vspace{-1.5mm}
    \label{tab:critic_network}
    \centering
    \begin{tabular}{l c c}
        \toprule[1.0pt]
        Setting & FID $\downarrow$ & Top-1 $\uparrow$ \\
        \hline
        separate copy & 0.101 & 0.504 \\
        shared w/o detach  & 0.122 & 0.501 \\
        shared w/ detach  & \textbf{0.099} & \textbf{0.505} \\
        \bottomrule[1.0pt]
    \end{tabular}
  \end{minipage}
  \hspace{.012\textwidth}
  \begin{minipage}{.3\textwidth}
    \setlength\tabcolsep{3pt}
    \subcaption{PPO Train Setting}
    \vspace{-1.5mm}
    \label{tab:ppo_epoch_batch}
    \centering
    \begin{tabular}{c c c c}
        \toprule[1.0pt]
        Epoch & Batch & FID $\downarrow$ & Top-1 $\uparrow$ \\
        \hline
        1 & 1 & 0.102 & 0.502 \\
        1 & 1/2 & 0.107 & 0.500 \\
        2 & 1 & \textbf{0.099} & \textbf{0.505} \\
        2 & 1/2 & 0.101 & 0.503 \\
        \bottomrule[1.0pt]
    \end{tabular}
  \end{minipage}\
\end{table}

Lacking ground-truth motion sequences prevents us from calculating motion-to-motion alignment scores for synthesized, unpaired examples. Consequently, we mitigate this by assigning each unpaired instance the batch mean motion-to-motion alignment (pseudo) score of paired samples, thereby ensuring that the reward scores for both paired and unpaired samples are at the same scale. We then assess the model's generalization capability with human evaluation. Our findings reveal that both the isolated motion-to-text alignment and the compositional alignment strategy enhance the model's generalizability. Interestingly, contradicting observations from the paired dataset analysis above, the model relying solely on motion-to-text alignment obtains more positive feedback from evaluators regarding its generalization capabilities, as shown in Figure \ref{fig:human_ablation}.

\parsection{PPO Configurations}
Additionally, we delve into the implications of varying configurations within the PPO algorithm. As shown in Table \ref{tab:critic_network}, sharing the initial layers of its neural architecture with the actor model has no adverse effect on overall performance but actively bolsters computational efficiency during the training phase. Note that the optimization of the critic network needs to be isolated from the shared layers via the detach operation to keep the learning of the actor model unaffected. We also conduct experiments varying the number of PPO epochs and the internal batch size, the latter being set as a fraction of the external batch size. The results in Table \ref{tab:ppo_epoch_batch} indicate that optimal performance is achieved with an epoch number of 2 and an internal batch size ratio of 1/2.

\section{Conclusion \& Future Work}

We present InstructMotion, a reinforcement learning-based framework for generalizable text-driven human motion synthesis. By formulating autoregressive generation as a Markov decision process and integrating it with curated reinforcement learning techniques inspired by RLHF, our method innovatively mitigates the key challenge of generalization, a limitation that characterizes contemporary text-to-motion methods. We design reward model underpinned by pre-trained text and motion encoders, sidestepping conventional preference collection requisites. Without necessitating the additional motion-text pairs, InstructMotion achieves remarkable improvement in generating semantically aligned novel human motions. Extensive evaluations on standard benchmarks like HumanML3D and KIT-ML demonstrate the superiority of our method, which yields not only improved quantitative evaluation metrics but also the leading human evaluation results given novel motion descriptions.

The generalization capability of InstructMotion is constrained by its reliance on the reward model, which are currently trained on limited paired data. Given that the reward model prioritize semantic alignment over the fine-grained details, a future direction involve exploiting the synthesized paired data (\eg generated by motion stitching) to train the reward model for better generalization capability.

{
\bibliographystyle{splncs04}
\bibliography{reference}
}
\end{document}